\documentclass{article}

     \usepackage[preprint,nonatbib]{neurips_2020}

\usepackage[utf8]{inputenc} 
\usepackage[T1]{fontenc}    
\usepackage{hyperref}       
\usepackage{url}            
\usepackage{booktabs}       
\usepackage{amsfonts}       
\usepackage{nicefrac}       
\usepackage{microtype}      
\usepackage{algorithm2e}

\usepackage{microtype}
\usepackage{graphicx}
\usepackage{subfigure}
\usepackage{booktabs} 
\usepackage{amsmath}
\usepackage{multirow}
\usepackage{makecell}
\usepackage[export]{adjustbox}

\usepackage{hyperref}

\author{
  Xingwen Zhang \\
  Ant Financial Services Group\\
  Sunnyvale, CA 94085\\
  \texttt{xingwen.zhang@antfin.com} \\
  \And
  Shuang Yang\thanks{Corresponding author} \\
  Ant Financial Services Group\\
  Sunnyvale, CA 94085\\
  \texttt{shuang.yang@antfin.com} \\
}

\title{Learning (Re-)Starting Solutions for Vehicle Routing Problems}

\begin{document}
\maketitle

\begin{abstract}
A key challenge in solving a combinatorial optimization problem is how to guide the agent (i.e., solver) to efficiently explore the enormous search space. Conventional approaches often rely on enumeration (e.g., exhaustive, random, or tabu search) or have to restrict the exploration to rather limited regions (e.g., a single path as in iterative algorithms). In this paper, we show it is possible to use machine learning to speedup the exploration. In particular, a value network is trained to evaluate solution candidates, which provides a useful structure (i.e., an approximate value surface) over the search space; this value network is then used to screen solutions to help a black-box optimization agent to initialize or restart so as to navigate through the search space towards desirable solutions. Experiments demonstrate that the proposed ``\emph{Learn to Restart}'' algorithm achieves promising results in solving \emph{Capacitated Vehicle Routing Problems} (CVRPs).

\end{abstract}

\section{Introduction}
One key challenge that makes solving combinatorial optimization problems particularly hard is the lack of an efficient exploration strategy. It's challenging as the number of potential candidates is daunting owing to the enormous search space. Conventional methods often employ search algorithms such as \emph{random search}, \emph{simulated annealing} \cite{Kirkpatrick1983} and \emph{tabu search} \cite{Glover1989} by using an iterative procedure, which starts at an initial solution (or a population of them) and moves from one solution to another by applying a set of pre-defined operators.
These approaches are far from satisfactory as they are either enumerative and thus only a small portion of the space can be explored within feasible budget of time and compute, or they have to restrict the exploration to very limited regions (e.g., a single / a few trajectories as in iterative algorithms). Is it possible to obtain an \emph{oracle} that can tell good from bad effortlessly, such that with its help the solving agent can navigate through the search space by avoiding bad solutions and only exploring the good ones? We show in this paper that the answer is affirmative.

In particular, we propose to train a value network to evaluate solution candidates, which provides a useful structure (i.e., an approximate value surface similar to that of a continuous optimization problem) over the search space. The trained value network is then used to screen solutions to help a \emph{black-box optimization agent} (BOA) to initialize or to restart so as to navigate through the search space towards desirable solutions. The approach is called ``\emph{Learn to Restart}'' (L2R), and it can work with any iterative optimization method without knowing how the optimization agent works internally.

We are interested in understanding how L2R helps combinatorial optimization either by reducing the number of runs of a BOA or by producing better solutions with the same number of runs. For this purpose, we test L2R on the \emph{Capacitated Vehicle Routing Problems} (CVRPs) \cite{Dantzig59}, a benchmark problem for learning-based optimization algorithms \cite{Bengio18, nazari2018reinforcement, kool2018attention, Chen19, Lu2020}. The CVRP is NP-hard, which could be solved 
using exact or heuristic methods \cite{fukasawa2006robust, golden2008vehicle, Kumar12, toth2014vehicle}. In a CVRP, we have a depot (denoted by 0) and a set of $N$ customers, each with demand $d_i \ (1 \leq i \leq N)$ to be fulfilled. The goal is to design a set of vehicle routes with minimal total traveling distance to serve all the customers. A vehicle always starts and ends at the depot, while the total demand of the customers served by a vehicle should not exceed its capacity (denoted by $C$). A detailed formulation is given in Supplemental 
Material. Our experiments demonstrate that L2R is able to outperform competitive baselines by a significant margin. We also explore the approach to solve combinatorial optimization by using an L2R oracle, a solving agent and a search policy together. In the literature, using machine learning to approximate the objective function has been studied before. For example, the STAGE method \cite{Boyan2000} trains a regression model to predict the objective function. It is worthwhile to note the differences between L2R and STAGE.
\begin{itemize}
     \item L2R is \emph{inductive} whereas STAGE is \emph{transductive}, i.e., L2R can be used to assess solutions for problem instances in general (i.e., not seen in training), while STAGE needs to be trained and tested on the same problem instance. 
     
    \item L2R is based on Siamese network and preferential learning, which is shown to be more effective than the regression-based methods such as STAGE.

    \item In testing, L2R is applied to randomly generated initial solutions to choose more promising starting points, while STAGE runs a separate optimization procedure over the approximate function to find new starting points.
\end{itemize}

    
    
    


Lastly, although in this paper our motivating task is CVRPs, the techniques developed are not specific to the formulation of CVRPs and equally applicable to other combinatorial optimization problems. 

\section{Overall Framework}
\begin{figure}[ht]
\centering
\begin{tabular}{ccc}
\includegraphics[width=0.31\linewidth]{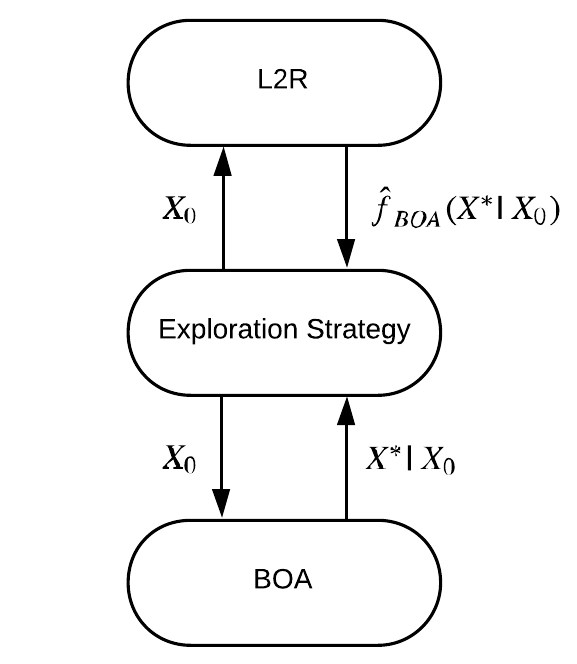} & & \includegraphics[width=0.6\linewidth]{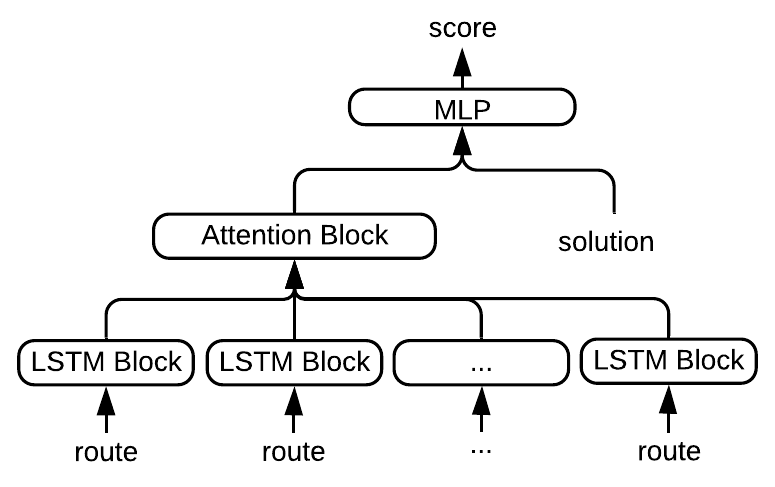}\\
~~~~~~~~(a) System diagram & & ~~~~~~~(b) Value network
\end{tabular}
\caption{System diagram and value network.}
\label{FigureFrameworkNetwork}
\end{figure}

Our overall framework, as given in Figure \ref{FigureFrameworkNetwork}(a), consists of three main components, that is, a black-box optimization agent (BOA), a Learn to Restart (L2R) module, and an exploration strategy that maintains a population of solutions (including the special case with a population of size one).

The BOA can be a classical optimizer for the problem under investigation (such as LKH3 \cite{helsgaun2017extension} for the CVRP) or a learnt optimizer (e.g. the RL-based agent \cite{Lu2020}). Our only assumption about the BOA is that it takes an initial solution of a problem instance, $X_0$, improves upon it and returns $X^{*}|X_0$, the solution optimized by the BOA that starts at $X_0$.

Given the solution space implicitly defined by the BOA, the L2R module, when given an initial solution, $X_0$, predicts the objective function value of $X^{*}|X_0$ using a value network, without actually running the BOA. The predicted value is denoted by \begin{flalign}
\hat{f}_{BOA}(X^{*} |\ X_0) \nonumber
\end{flalign}
in Figure \ref{FigureFrameworkNetwork}(a). For a set of initial solutions, $\hat{f}_{BOA}$ can be used to predict which of them will lead to a better resultant solution produced by the BOA. If a model can be trained offline using problem instances sampled from the same distribution as the test instances, L2R could potentially help reduce the number of BOA runs or produce better solutions with the same number of runs.

Lastly, the exploration strategy decides how the BOA and L2R model would interact with each other. When there is a single individual in the population, L2R can be used to select better offspring of the individual. For a population of more than one individual, L2R, besides its usage in offspring selection as before, can also be used to maintain the population size, as well as to select individuals out of the population that would then be optimized by the BOA. Specifically, when the size of the population exceeds a desired number, $I$, L2R is used to shrink the population to the desired size by keeping only the $I$ best individuals as predicted by the value network. For next iteration, L2R is used to select the best individual out of the population to be evaluated by the BOA.

\section{Learn to Restart}
The motivation of L2R is to predict the final distance we would get from running the BOA on an initial solution without actually running the agent. Since sampling a set of initial solutions is often inexpensive (e.g., for CVRP, solutions could be  generated by randomly ordering the customers and serving them sequentially and adding new vehicles when necessary), we can then generate a set of initial solutions, predict the corresponding distances that would be achieved by the BOA, and only improve those that are promising as per the oracle predictor.

The function $\hat{f}_{BOA}$ is approximated by a value neural network, which is a universal function approximator \cite{Cybenko1989, Leshno1993}. The solution spaces of typical combinatorial optimization problems are often highly irregular (e.g., discrete, non-continuous and non-differentiable) due to the existence of complex constraints (e.g., integer constraints). It is thus reasonable for us to design a value network to approximate the function, taking advantage of automatic feature engineering capability of a neural network.

Below we describe in details training data generation, the structure of the network, and how the network is trained and applied.

\subsection{Data Generation}
To produce training samples, we randomly sample a large number of problem instances. For each problem instance, two random initial solutions are sampled and then improved by the BOA, and the resultant distances are used to label the initial solutions. In particular, training labels are generated based on whether one solution in a pair is better than the other by a margin (denoted as $\epsilon$). It is worthwhile to point out that this data generation process can be done offline in parallel. Furthermore, it can be done using CPUs only if the BOA needs no GPU.

\subsection{Training Value Network}
For each pair of initial solutions of the same training problem instance, a Siamese neural network \cite{Bromley1993} is trained to predict which initial solution will lead to a smaller distance returned by the BOA. Half of the neural network is given in Figure \ref{FigureFrameworkNetwork}(b) (the other half is identical and shares the same weights), where the LSTM blocks share the same weights. Specifically, each route in a solution is first embedded using an LSTM block, which consists of a bi-directional LSTM layer and a subsequent fully connected layer. A route is represented as a sequence of customers, where each element of the sequence contains customer-specific features such as its location and demand, as well as features about the neighboring nodes such as the distance from the previous node and the distance to the next node. The LSTM embedding of each route is then concatenated and fed into an attention \cite{Vaswani17} block, which consists of a number of attention stacks. Solution-specific features including the sum and standard deviation of the distances of the routes are appended to the embedding produced by the attention block, which are then fed to an MLP to produce a score for the solution. 

The scores of a pair of solutions are transformed to two logit values, one for each. Given the logit values and the label associated with each pair of solutions, a cross-entropy loss is defined as usual,
\begin{flalign}
loss = -\sum_{s \in S} (y_s * log(p_s) + (1 - y_s) * log(1 - p_s)), \nonumber
\end{flalign}
where $S$ is the set of training samples, and $y_s$ and $p_s$ are the label and predicted probability of having label 1 for a sample $s$, respectively. The network is trained using ADAM with a learning rate of 0.0005.

In our design we use a Siamese network to evaluate pairs of solutions of the same problem instance. We believe that it makes more sense to compare solutions of the same instance, than of different instances, by eliminating the impact of problem-specific features. Furthermore, when integrated into our overall optimization framework, an L2R model is used only for pairs of solutions of the problem instance under consideration. It is thus reasonable for the network to be tailor made for its eventual usage.

\subsection{Using Value Network}
A trained L2R network can be used in a number of ways in our overall optimization framework. We propose and experiment with two ways of integrating L2R into the optimization framework. 

Firstly, upon the convergence of the BOA, we generate a new set of initial solutions either through fully random permutation of the customers or by destructing and reconstructing $R$ existing routes in the solution returned by the BOA. These new initial solutions are then evaluated by the L2R model, and only the best of them is given to the BOA for further improvement. The pseudo-code is given Algorithm \ref{AlgorithmSequential}.
\begin{algorithm}
\DontPrintSemicolon
\SetAlgoLined
    $X^{*}$ = None \;
    $d^{*}$ = $\infty$ \;
    generate a random solution $X^{*}_{0}$ \;
    \ForEach{$t \in \{1,\dots,T\}$} {
        generate $O$ individuals by random permutation or destructing and reconstructing $R$ routes \;
        predict the best of the $O$ individuals, denoted by $X_0$, using trained Siamese network \;
        compute $X^{*}_{0}$ by invoking the BOA with $X_0$ as the starting solution \;
        calculate the total distance traveled, $d^{*}_{0}$, for $X^{*}_{0}$ \;
        \If{$d^{*}_{0} <  d^{*}$} {
            $X^{*} = X^{*}_{0}$ \;
            $d^{*} = d^{*}_{0}$
        }
    }
\Return $X^{*}$
\caption{Using trained Siamese network in a sequential setting}
\label{AlgorithmSequential}
\end{algorithm}

Secondly, in a population-based approach, we maintain a population of $I$ individuals. We use L2R to pick one of them, remove it from the population and improve it by the BOA. Upon convergence we produce $O$ offspring of the solution returned, and L2R is again applied to keep top $I$ out of the $I+O-1$ candidates. Thus, the population size is kept at $I$ for each iteration of the BOA.
The pseudo-code is given Algorithm \ref{AlgorithmPopulation}.
\begin{algorithm}
\DontPrintSemicolon
\SetAlgoLined
    $X^{*}$ = None \;
    $d^{*}$ = $\infty$ \;
    generate $\Psi_X$, a set of $I$ random initial solutions \;
    \ForEach{$t \in \{1,\dots,T\}$} {
        predict the best of the $I$ individuals, denoted by $X_0$, using trained Siamese network \;
        compute $X^{*}_{0}$ by invoking the BOA with $X_0$ as the starting solution \;
        calculate the total distance traveled, $d^{*}_{0}$, for $X^{*}_{0}$ \;
        \If{$d^{*}_{0} <  d^{*}$} {
            $X^{*} = X^{*}_{0}$ \;
            $d^{*} = d^{*}_{0}$
        }
        generate $O$ offspring of $X^{*}_{0}$ by destructing and reconstructing $R$ routes \;
        add the offspring to $\Psi_X$, and remove $X_0$ from $\Psi_X$ \;
        shrink $\Psi_X$ by keeping only the $I$ best individuals predicted by trained Siamese network \;
    }
\Return $X^{*}$
\caption{Using trained Siamese network in a population-based setting}
\label{AlgorithmPopulation}
\end{algorithm}

\section{Experiments}
Our experiments are designed in order to investigate the following questions.
\begin{itemize}
    \item
    Is the value network able to learn useful information from the training samples? In particular, we are interested in its out-of-sample prediction accuracy.
    
    \item
    What is the effect of a learnt L2R model on the final distances achieved?
    
    \item
    Can an L2R model be generalized to a different distribution of initial solutions?

    \item
    What is the impact of L2R when applied to an evolving population of solutions?
\end{itemize}

The training and test data used in our experiments are generated following the same protocol as \cite{nazari2018reinforcement, kool2018attention, Chen19, Lu2020}. Specifically, we experiment with CVRP-20, CVRP-50 and CVRP-100, each with 20, 50 and 100 customers, respectively. The capacity of the depot is 30, 40 and 50, respectively. For each training / test case, the location of the depot and those of the customers are uniformly sampled from a unit square (with (0.0, 0.0) and (1.0, 1.0) as the lower left and upper right corners, respectively), and the demand of each customer is uniformly sampled from the set $\{1, 2, \dots, 9\}$.

The LSTM block of the value network consists of a bi-directional LSTM layer with 64 hidden units, followed by a fully connected layer with 64 output units. The attention block consists of an attention layer with 8 heads and 128 units each, followed by a fully connected layer with 64 output units. The attention embedding is concatenated with solution-specific features, producing a score after two fully connected layers of size 128 and 1, respectively.

Lastly, the BOA iteratively applies a set of RL-learnt \cite{Lu2020} improvement operators, as given in Table \ref{TableMostlyUsedOperators} until no further improvement is possible.
\begin{table*}[ht]
\begin{center}
\caption{Improvement operators used}
\vspace{10pt}
\begin{tabular}{c|c|c}
\hline
Class & Name & Details \\
\hline
\multirow{3}{*}{Intra-route} 
& 2-Opt & \makecell{Remove two edges and \\ reconnect their endpoints} \\
\cline{2-3}
\cline{2-3}
& Relocate(1) & \makecell{Move a customer in the route to \\ a new location} \\
\hline
\multirow{4}{*}{Inter-route} & Cross(2) & Exchange the tails of two routes \\
\cline{2-3}
& Symmetric-exchange(2) & \makecell{Exchange segments of length $m$ \\ ( $m=1$) between two routes} \\
\cline{2-3}
& Relocate(2) & \makecell{Move a segment of length $m$ \\ ( $m=1$) from a route to another} \\
\hline
\end{tabular}
\label{TableMostlyUsedOperators}
\end{center}
\end{table*}

\subsection{L2R as a Learner}
We generated a test set of 150000 pairs of random initial solutions. After computing their final distances by applying the BOA, a pair was discarded if the difference between the final distances was less than 0.01 (that is, $\epsilon=0.01$), otherwise a label of 0 or 1 was generated depending on which solution was better. The training data was generated in the same way.

To measure the quality of learning, we calculate the out-of-sample prediction accuracy and AUC. Our experimental results showed that the learning capability of the L2R model is relatively weak. For example, after being trained with 75 millions CVRP-20 instances (20 epochs over 3750000 random instances), its test accuracy is about 53.41\%, while its AUC is about 0.56. Similar metrics were observed for CVRP-50 and CVRP-100 (see Table \ref{TableTraining} for details).
\begin{table}[ht]
\centering
\caption{Training accuracy and AUC for CVRP}
\vspace{10pt}
\begin{tabular}{c|c|c|c}
\hline
 & CVRP-20    &  CVRP-50  & CVRP-100  \\
\hline
Accuracy & 53.41\% &  54.11\%  & 52.66\%  \\
\hline
AUC & 0.5559    &  0.5611  & 0.5511 \\
\hline
\end{tabular}
\label{TableTraining}
\end{table}

Furthermore, to investigate how the complexity of the BOA affects the test accuracy of the L2R model, we ran a set of experiments on CVRP-20 with different BOA's. For this purpose, we introduced a hyper-parameter $M$ that explicitly controls the maximum number of optimization cycles (where each cycle corresponds to a sequential application of the operators listed in Table \ref{TableMostlyUsedOperators}) used by the BOA. Since the underlying algorithm and data set stay the same, it is intuitive that the complexity increases as $M$ increases.
As shown in Table \ref{TableTrainingComplexity}, the test accuracy and AUC generally decrease as the complexity increases. This is reasonable, since it is harder to predict the outcome of the optimization as it becomes more sophisticated.
\begin{table}[ht]
\centering
\caption{Training accuracy and AUC for BOA's of varying complexity}
\vspace{10pt}
\begin{tabular}{c|c|c|c|c|c}
\hline
$M$ & 1    &  2 & 3 & 4 & 5  \\
\hline
Accuracy &  59.23\% &   54.86\%  &  51.43\% & 52.52\% & 51.84\% \\
\hline
AUC &  0.6318   &   0.5775  &   0.5323 & 0.5498 & 0.5407 \\
\hline
\end{tabular}
\label{TableTrainingComplexity}
\end{table}

We also experimented with a regression-based L2R model that tries to directly predict the value of an initial solution without using the Siamese network. The network structure is identical to the half of the Siamese network. Our experimental results showed that, when applied to a pair of solutions of the same problem, the distance predicted by the regression model failed to differentiate the better from the worse.

\subsection{L2R Reduces Final Distances} \label{SectionFullyRandom}
\begin{figure}[ht]
\centering
\begin{tabular}{ccc}
\includegraphics[width=0.455\linewidth]{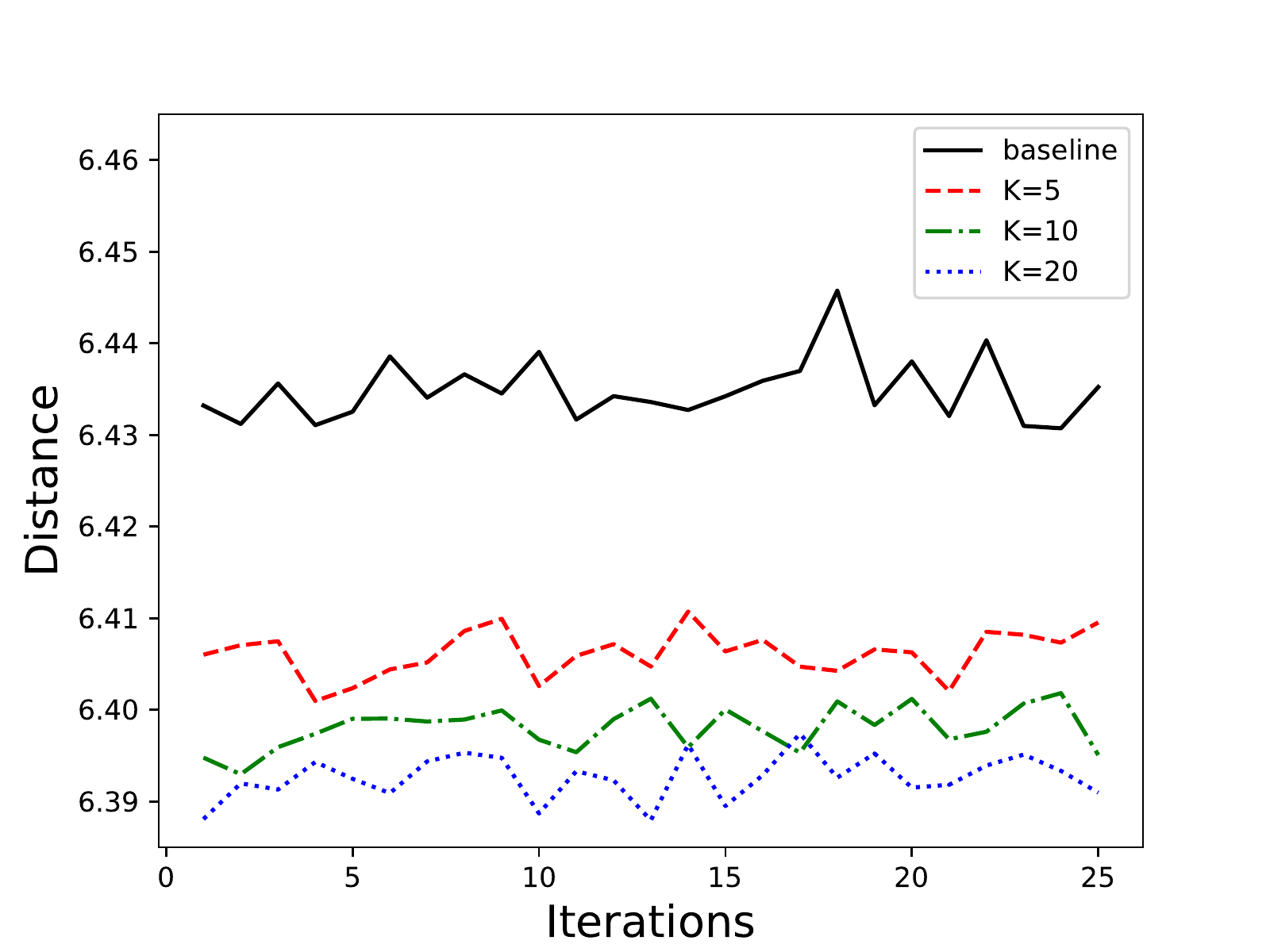} & & \includegraphics[width=0.455\linewidth]{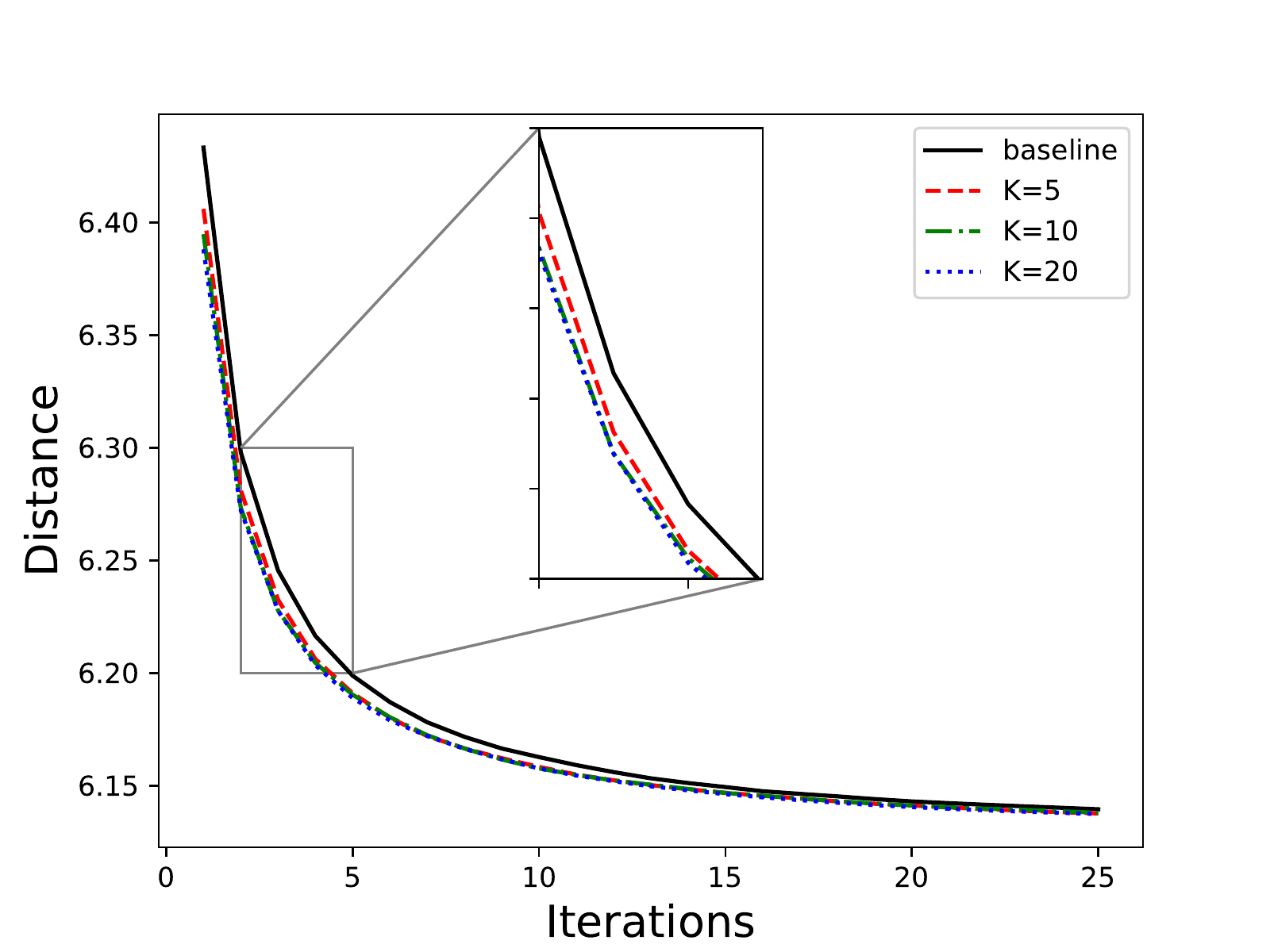}\\
~~~~~~~~(a) Avg distance for each iteration & & ~~~~~~~(b) Avg best distance after each iteration
\end{tabular}
\caption{CVRP-20 with fully random initial solutions.}
\label{FigureFullyRandom}
\end{figure}

To investigate the impact of the learnt model on the final solution quality, we ran a set of experiments and compared the average distance achieved in baseline and test settings. The baseline samples a random problem instance, generates a fully random initial solution and improves it until convergence, and regenerates another fully random initial solution and improves it until reaching the maximum number of BOA iterations. A test experiment is run on the same sequence of problem instances as the baseline (for ease of comparison), and for each BOA iteration it generates $K$ ($K \in \{5, 10, 20\}$) random initial solutions and applies L2R model to pick the most promising one and improves it until convergence. Figure \ref{FigureFullyRandom}(a) plots the average distance over 10000 CVRP-20 problem instances for each BOA iteration, from which we observe that L2R, as a weak learner, helps reduce the distance for each iteration. Furthermore, its impact becomes more profound as $K$ grows. Figure \ref{FigureFullyRandom}(b) illustrates the average best distance (that is, the best distance achieved until current iteration) for the same experiments, which again shows that L2R facilitates producing better solutions, although the impact becomes less significant as the number of BOA iterations increases.

\begin{figure}[ht]
\centering
\begin{tabular}{ccc}
\includegraphics[width=0.45\linewidth]{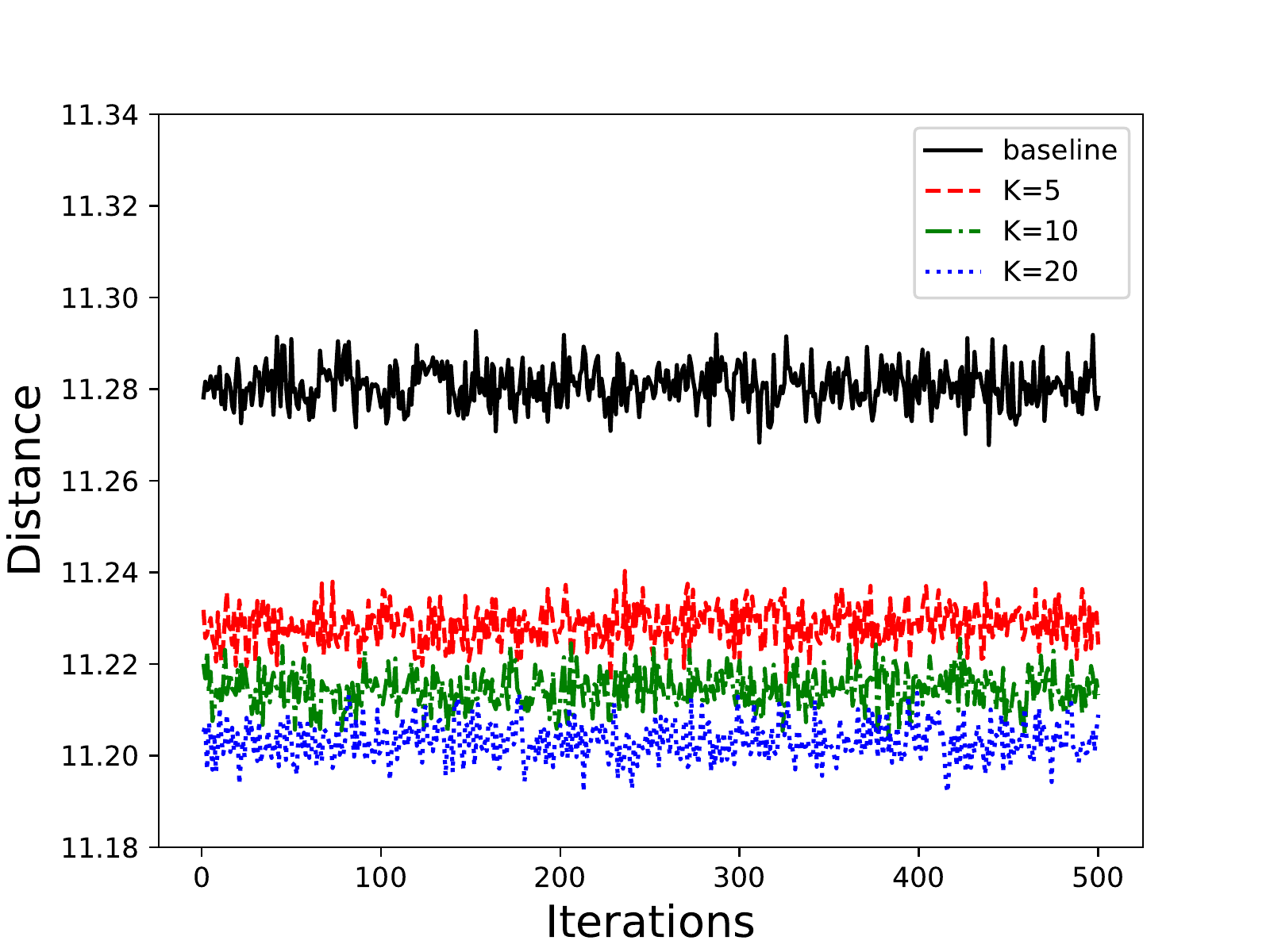} & & \includegraphics[width=0.45\linewidth]{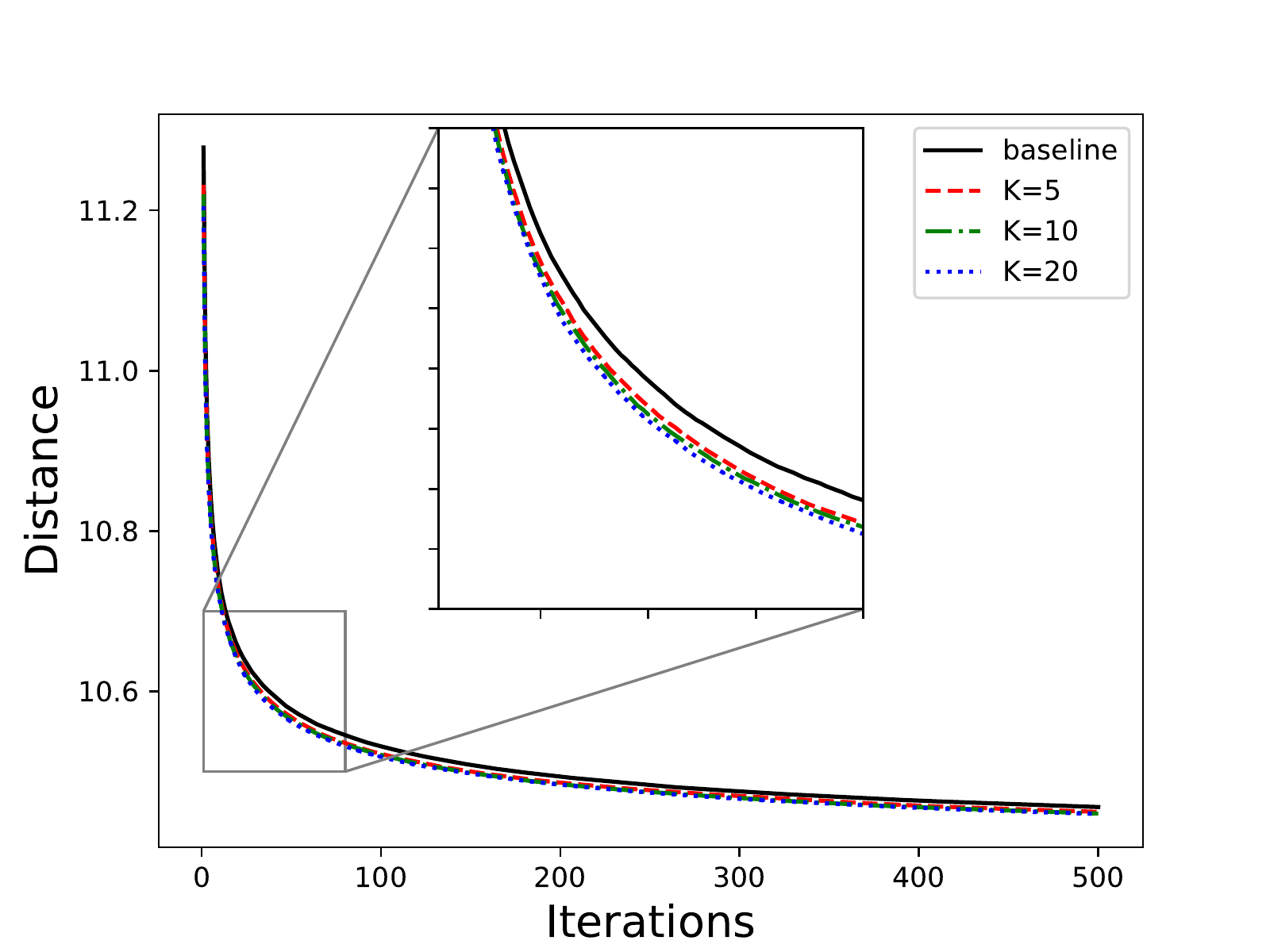}\\
~~~~~~~~(a) Avg distance for each iteration & & ~~~~~~~(b) Avg best distance after each iteration
\end{tabular}
\caption{CVRP-50 with fully random initial solutions.}
\label{FigureFullyRandom50}
\end{figure}

\begin{figure}[ht]
\centering
\begin{tabular}{ccc}
\includegraphics[width=0.455\linewidth]{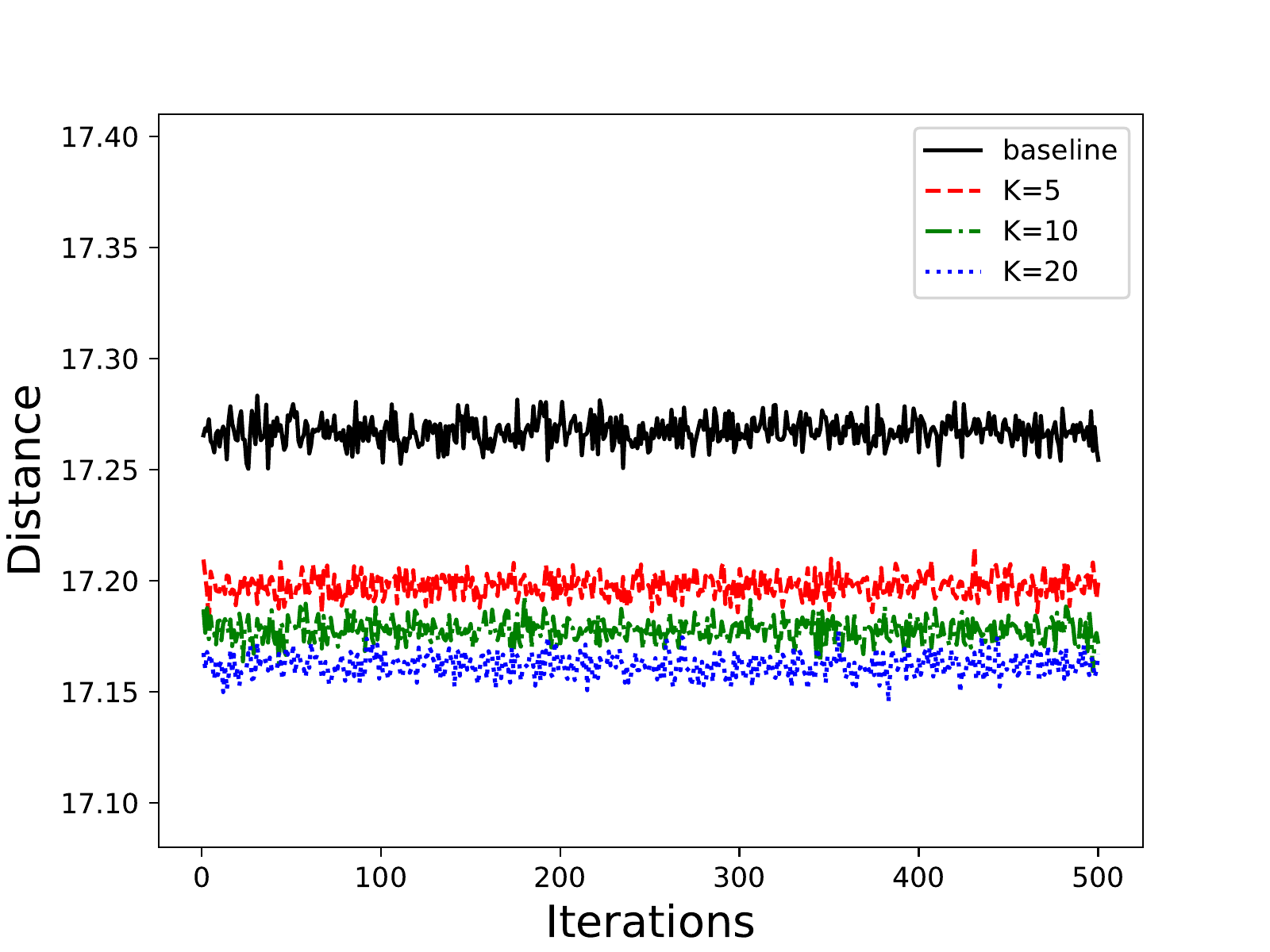} & & \includegraphics[width=0.455\linewidth]{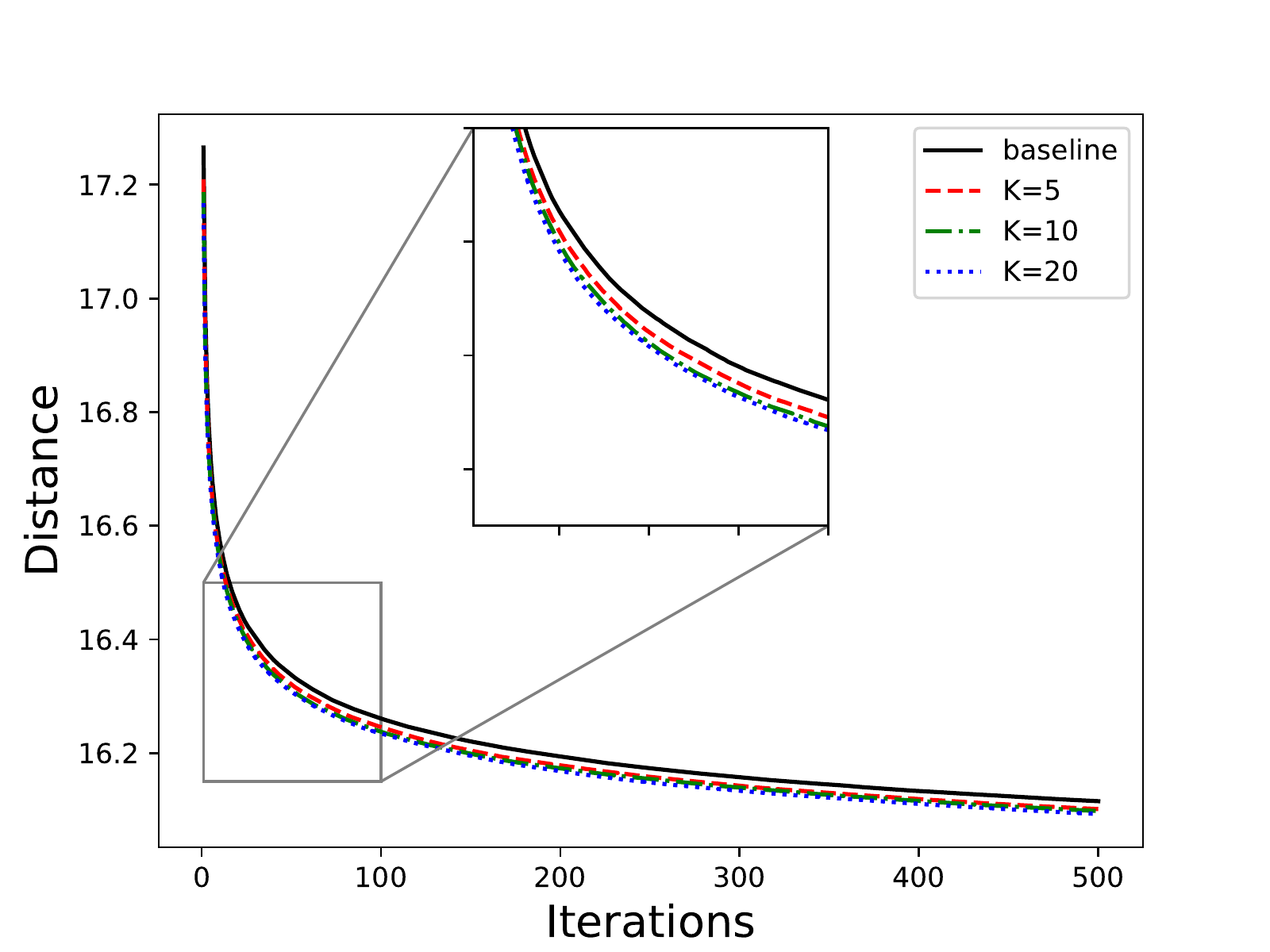}\\
~~~~~~~~(a) Avg distance for each iteration & & ~~~~~~~(b) Avg best distance after each iteration
\end{tabular}
\caption{CVRP-100 with fully random initial solutions.}
\label{FigureFullyRandom100}
\end{figure}

Lastly, as shown in Figure \ref{FigureFullyRandom50} and \ref{FigureFullyRandom100}, L2R improves the solution quality for CVRP-50 and CVRP-100 as well.

\subsection{L2R Generalized to Different Distributions of Initial Solutions} \label{SectionPartiallyPerturbed}
\begin{figure}[ht]
\centering
\begin{tabular}{ccc}
\includegraphics[width=0.455\linewidth]{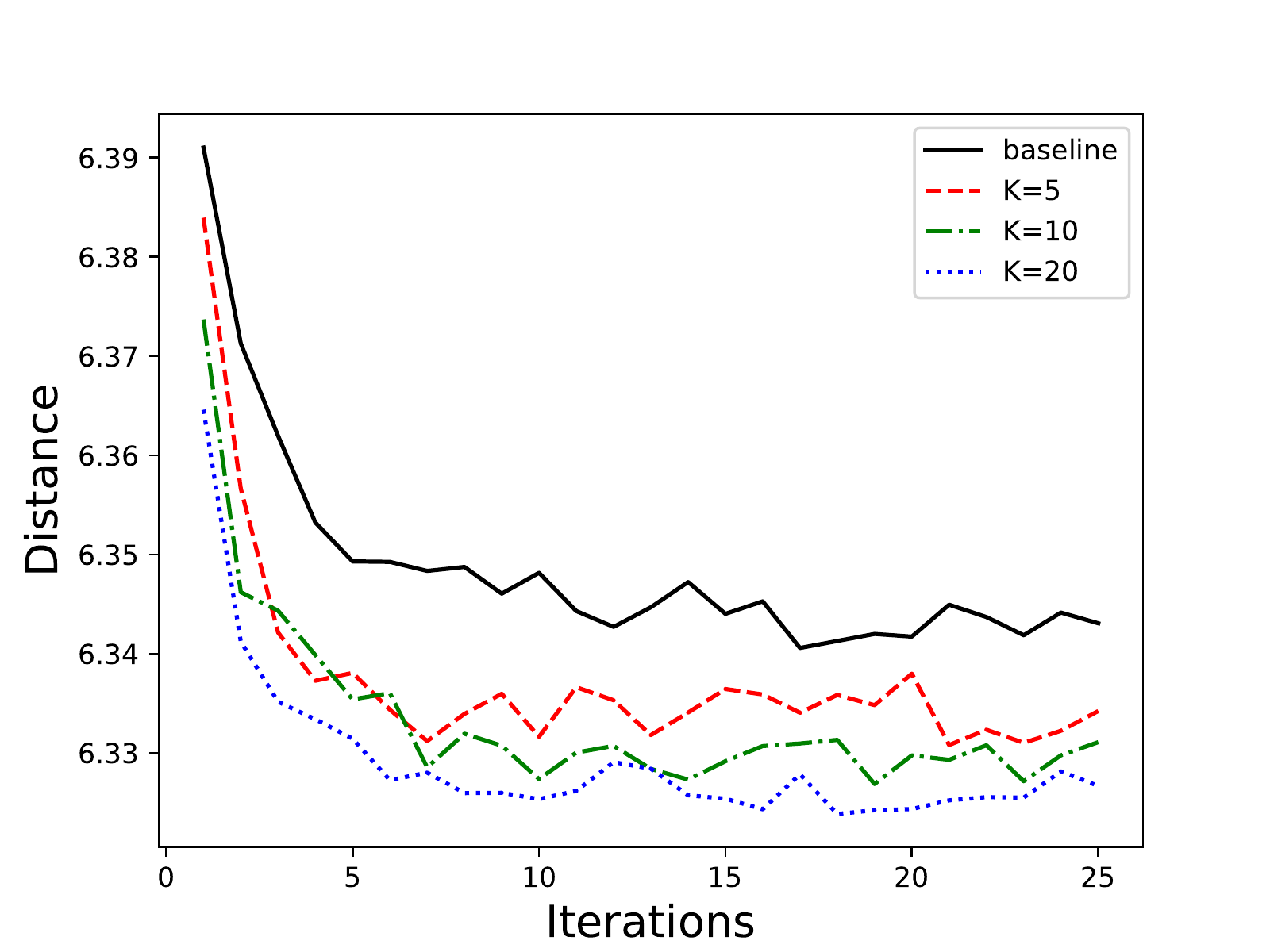} & & \includegraphics[width=0.455\linewidth]{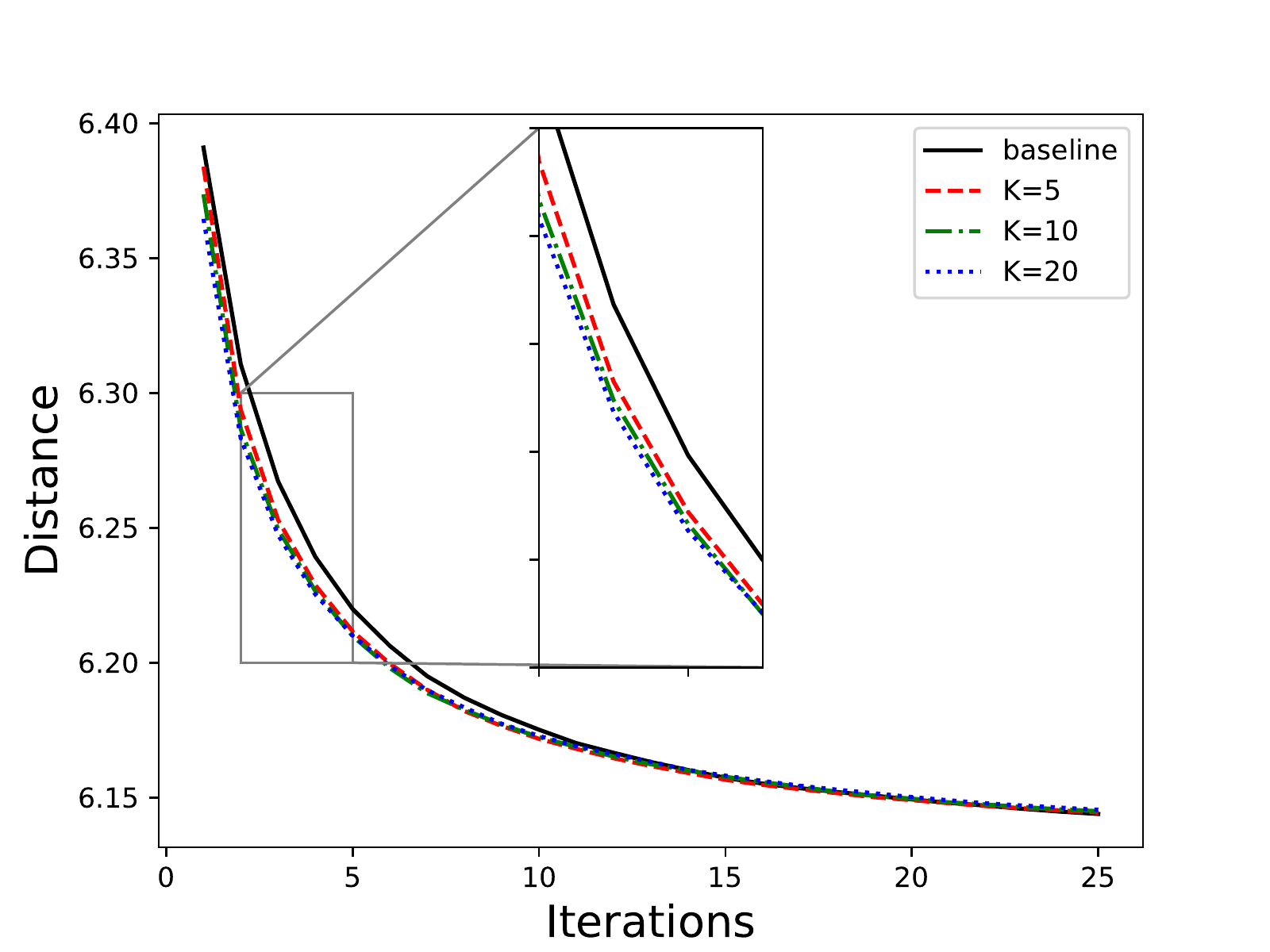}\\
~~~~~~~~(a) Avg distance for each iteration & & ~~~~~~~(b) Avg best distance after each iteration
\end{tabular}
\caption{CVRP-20 with partially perturbed initial solutions.}
\label{FigurePartiallYPerturbed}
\end{figure}

To analyze the behavior of L2R when the initial solutions are produced in a way different from the training data, we ran another set of experiments where the baseline now uses a history-dependent perturbation procedure. Specifically, with a solution returned by the BOA the perturbation procedure randomly picks $R=2$ routes (instead of all routes of the solution as done in Section \ref{SectionFullyRandom}) and shuffles the customers of the chosen routes. The test experiment uses the same perturbation procedure as the baseline, but produces $K$ partially perturbed initial solutions, instead of one. Again, the BOA is applied to the most promising initial solution as predicted by L2R. Figure \ref{FigurePartiallYPerturbed}(a) and \ref{FigurePartiallYPerturbed}(b) shows that L2R helps improve the solution quality again. It is worthwhile to point out that the average distance in Figure \ref{FigurePartiallYPerturbed}(a) generally decreases with BOA iterations, because the initial solutions are gradually becoming better.
Moreover, L2R is shown to improve the solution quality for CVRP-50 and CVRP-100 as well, when generalized to partially perturbed initial solutions (we omit the plots due to limited space).

Lastly, due to space limit the experimental results and analysis for population-based exploration strategies are given in the Supplementary Material.

\section{Conclusion}
In this paper, we proposed a Learn to Restart model that predicts, for any initial solution, the resultant solution quality produced by a black-box optimization agent. A Siamese value network was trained and integrated with a number of exploration strategies. The experimental results showed that L2R outperformed the baseline with no learning.

Nevertheless, our results also indicate that a direction of future work to investigate more ways of integrating L2R with exploration strategies, since sometimes loss of diversity may lead to a deterioration of solution quality. Moreover, active learning in L2R is also worth exploration, since when we optimize for a particular problem instance multiple initial solutions are actually evaluated by the BOA, which can be used to adjust the weights of the value network for that specific problem instance.

\nocite{Toth02}
\bibliographystyle{plain}
\bibliography{neurips_2020}

\end{document}


\maketitle

\section{Capacitated Vehicle Routing Problems}
 Let $V = \{0, 1, \dots, N\}$ denote the set of nodes containing the depot and the customers, $M$ be the number of vehicles available\footnote{It is assumed, w.l.o.g., that $M=N$ for the CVRP we consider.}, and $c_{i,j}$ be the traveling distance from node $i$ to $j$. An integer programming formulation of the CVRP is given as follows \cite{Toth02},
\begin{flalign}
& \min_{x_{i,j}}   \sum_{i \in V}\sum_{j \in V}{c_{i,j}x_{i,j}} \label{VRPObjective} \\
& s.t.  \nonumber \\
&  \sum\limits_{i \in V}{x_{i,j}} = \sum\limits_{j \in V}{x_{i,j}} = 1, \ \forall i, j \in V \setminus \{0\} \label{ConstraintIn} \\
&  \sum\limits_{i \in V}{x_{i,0}} = \sum\limits_{j \in V}{x_{0,j}} = M, \label{ConstraintDepotIn} \\
&  u_i - u_j + C x_{i,j} \leq C - d_j, \ \forall i\neq j, \ d_i + d_j \leq C  \label{ConstraintLoadUpdate} \\
&  d_i \leq u_i \leq C, \ \forall i \in V \setminus \{0\} \label{ConstraintLoadRange} \\
& x_{i,j} \in \{0,1\}, \forall i , j \in V. 
\end{flalign}
where 
constraints (\ref{ConstraintLoadUpdate}) and (\ref{ConstraintLoadRange}) impose the vehicle capacity requirements.

\section{L2R Integrated with Different Exploration Strategies}
In evolutionary algorithms, we often maintain an evolving population of solutions, and apply an optimization procedure to a subset of the population. We are interested in whether L2R is helpful in selecting the right individuals for the optimization procedure to work on. For this purpose we modified the experimental settings by maintaining an population of 20 individuals, and for each BOA iteration we apply L2R to rank the 20 individuals first. The top individual is chosen and then improved by the BOA. Upon convergence, $K$ offspring are generated by partial perturbation and the best offspring as predicted by L2R replaces their parent in the population. Thus, we always maintain a population of the same size.

\begin{figure}[ht]
\centering
\begin{tabular}{c}
\includegraphics[width=0.9\linewidth]{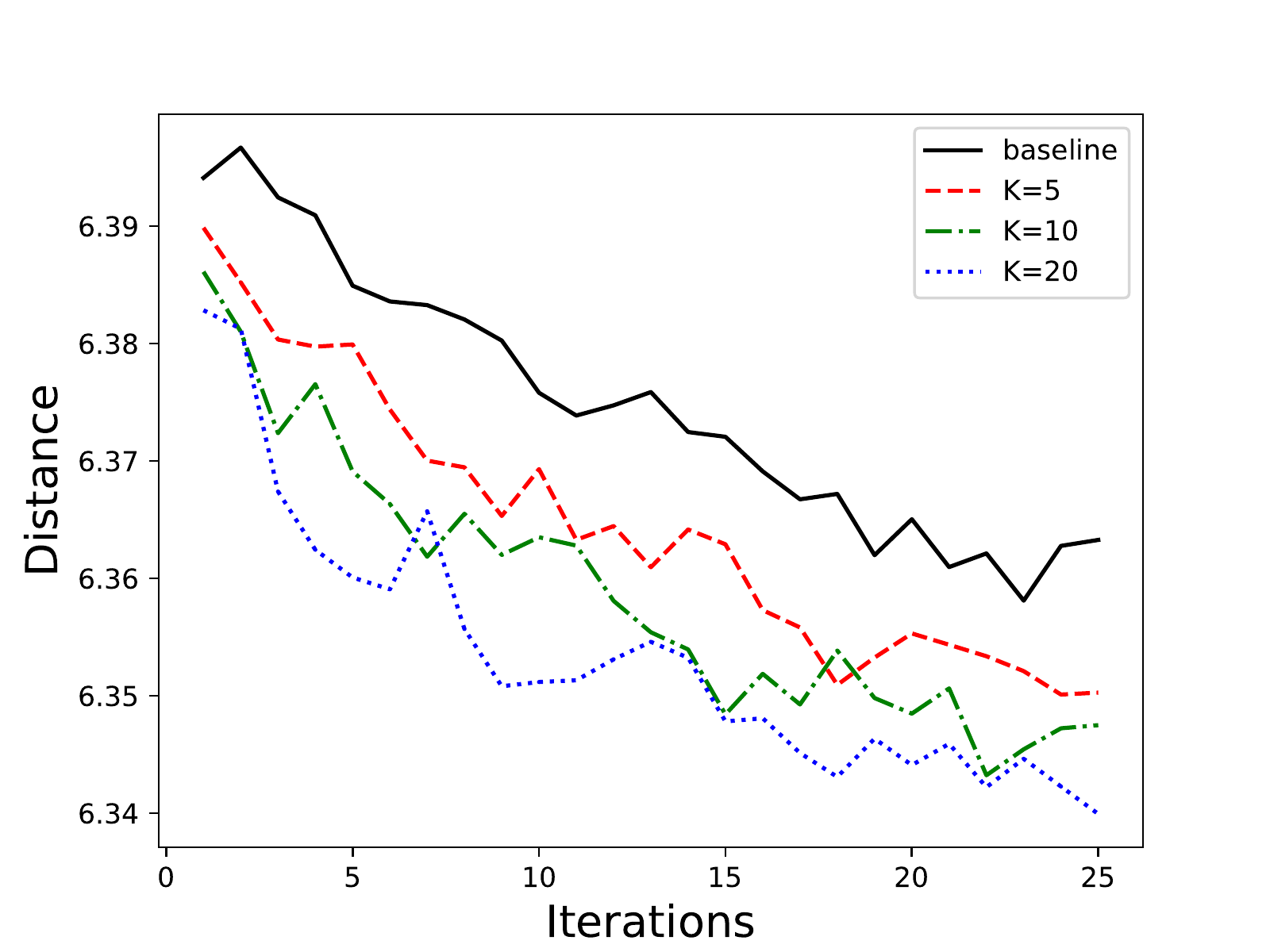}\\
(a) Average distance for each iteration  \\
\includegraphics[width=0.9\linewidth]{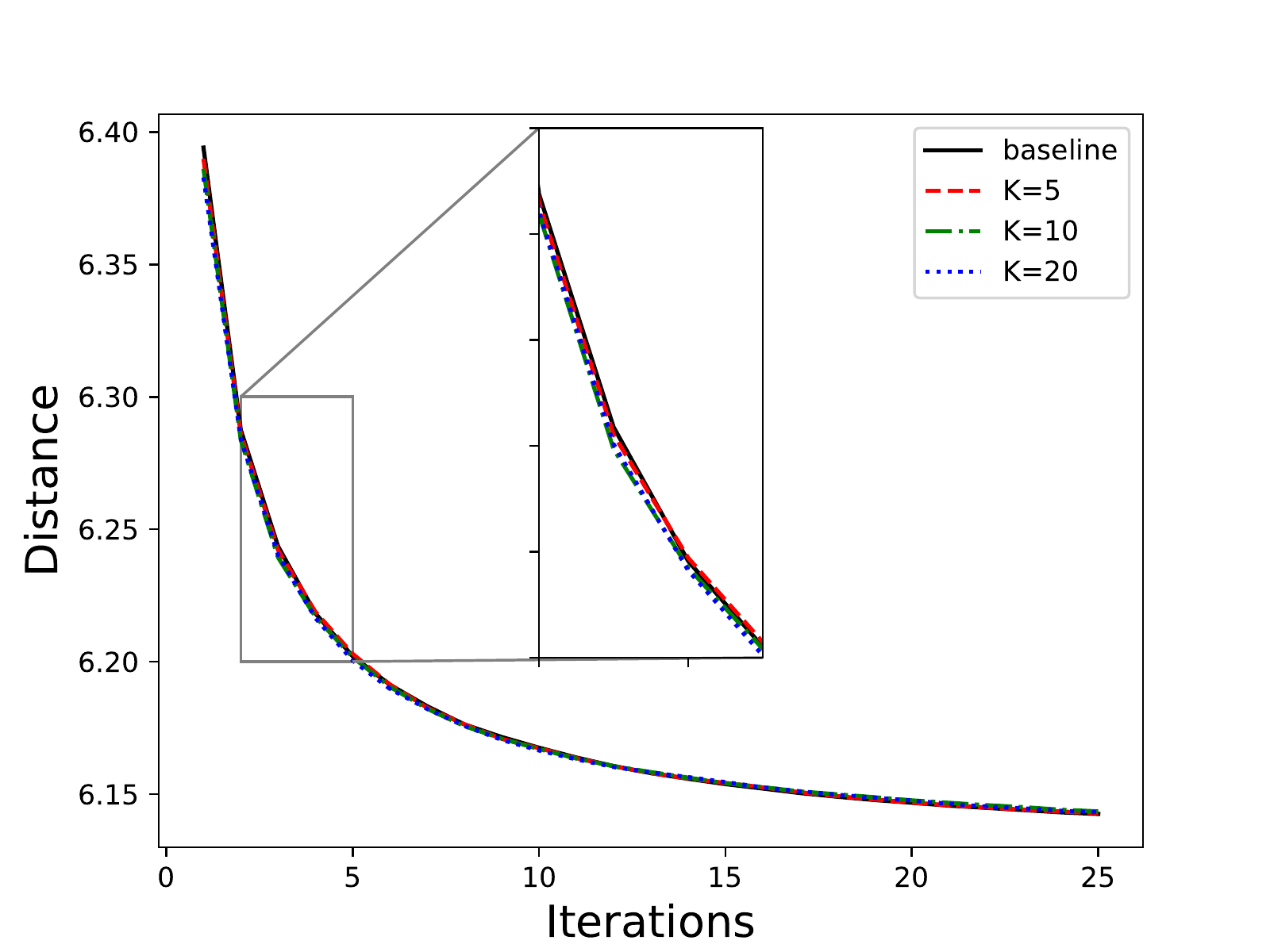} \\
(b) Average best distance after each iteration
\end{tabular}
\caption{CVRP-20: L2R chooses the best offspring.}
\label{FigureP20}
\end{figure}

\begin{figure}[ht]
\centering
\begin{tabular}{c}
\includegraphics[width=0.9\linewidth]{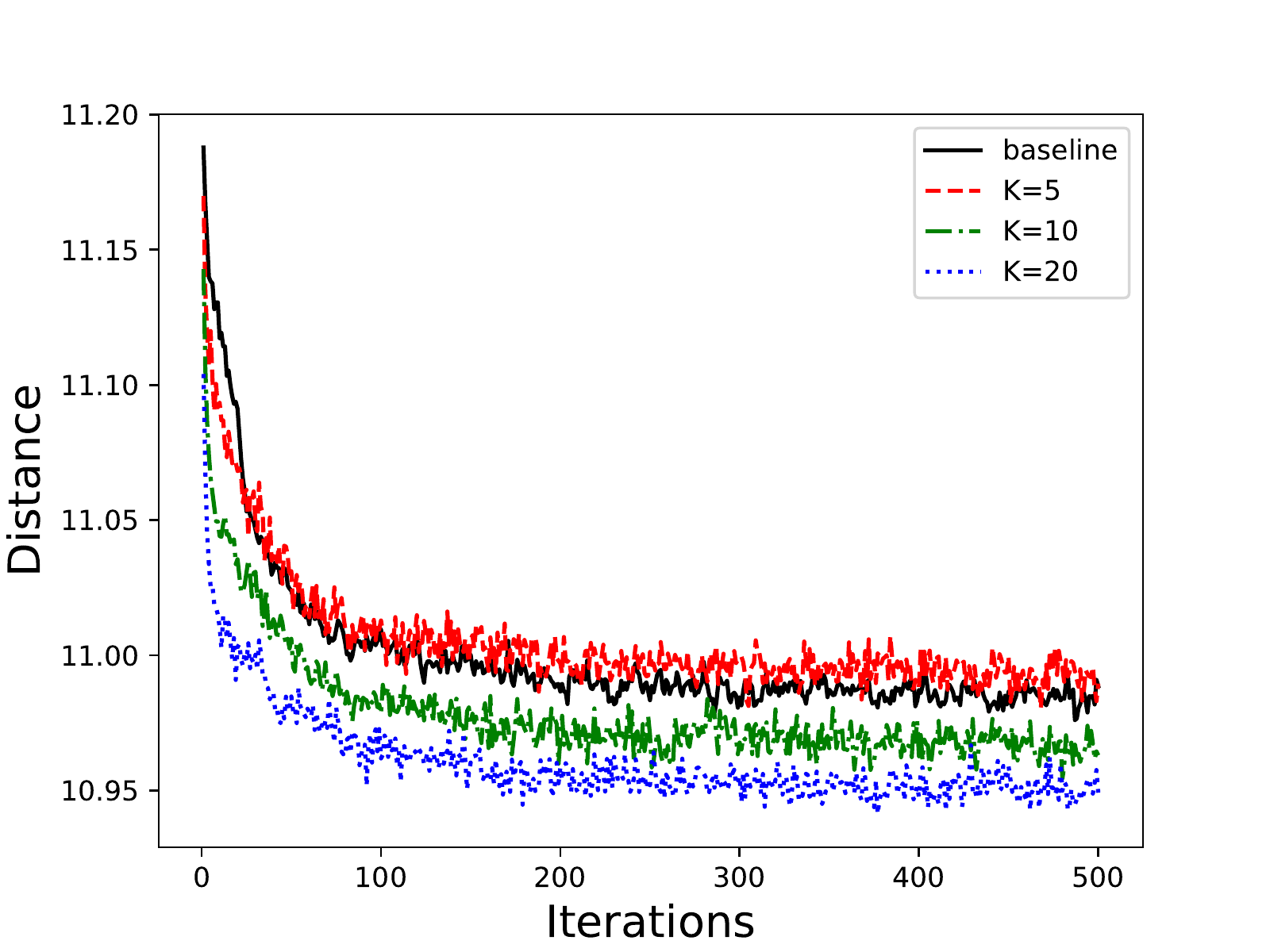}\\
(a) Average distance for each iteration  \\
\includegraphics[width=0.9\linewidth]{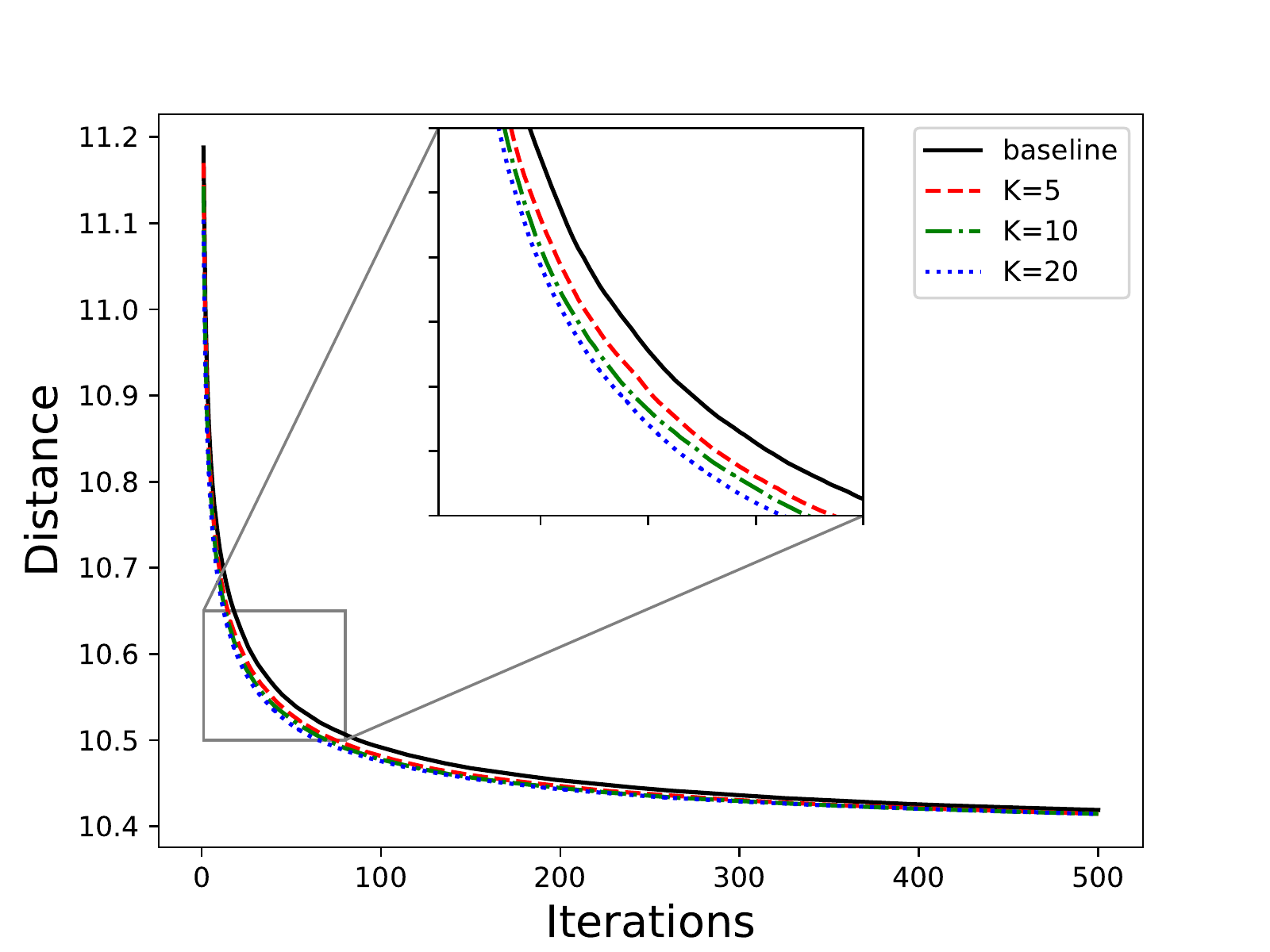} \\
(b) Average best distance after each iteration
\end{tabular}
\caption{CVRP-50: L2R chooses the best offspring.}
\label{FigureN50P20}
\end{figure}

\begin{figure}[ht]
\centering
\begin{tabular}{c}
\includegraphics[width=0.9\linewidth]{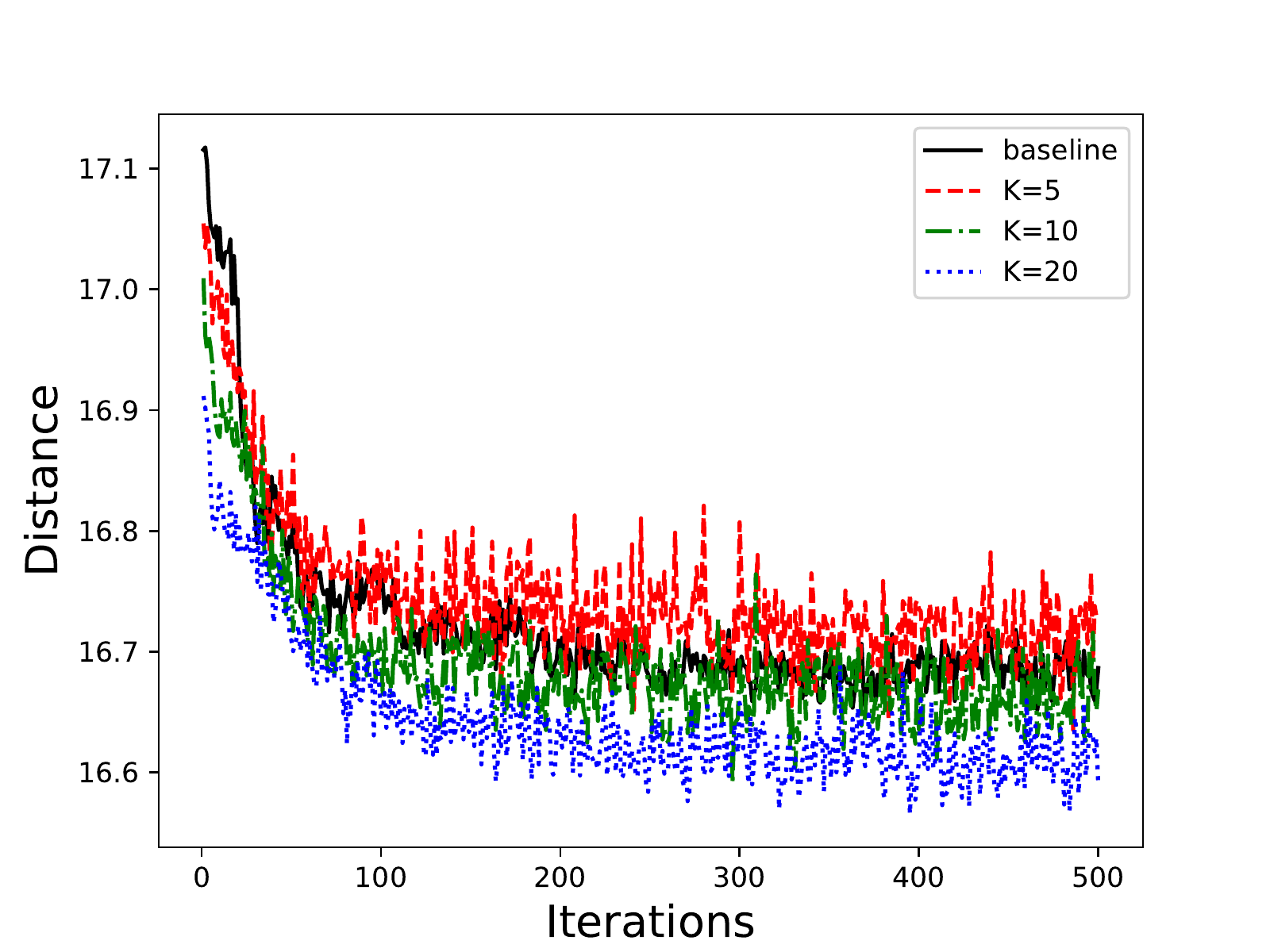}\\
(a) Average distance for each iteration  \\
\includegraphics[width=0.9\linewidth]{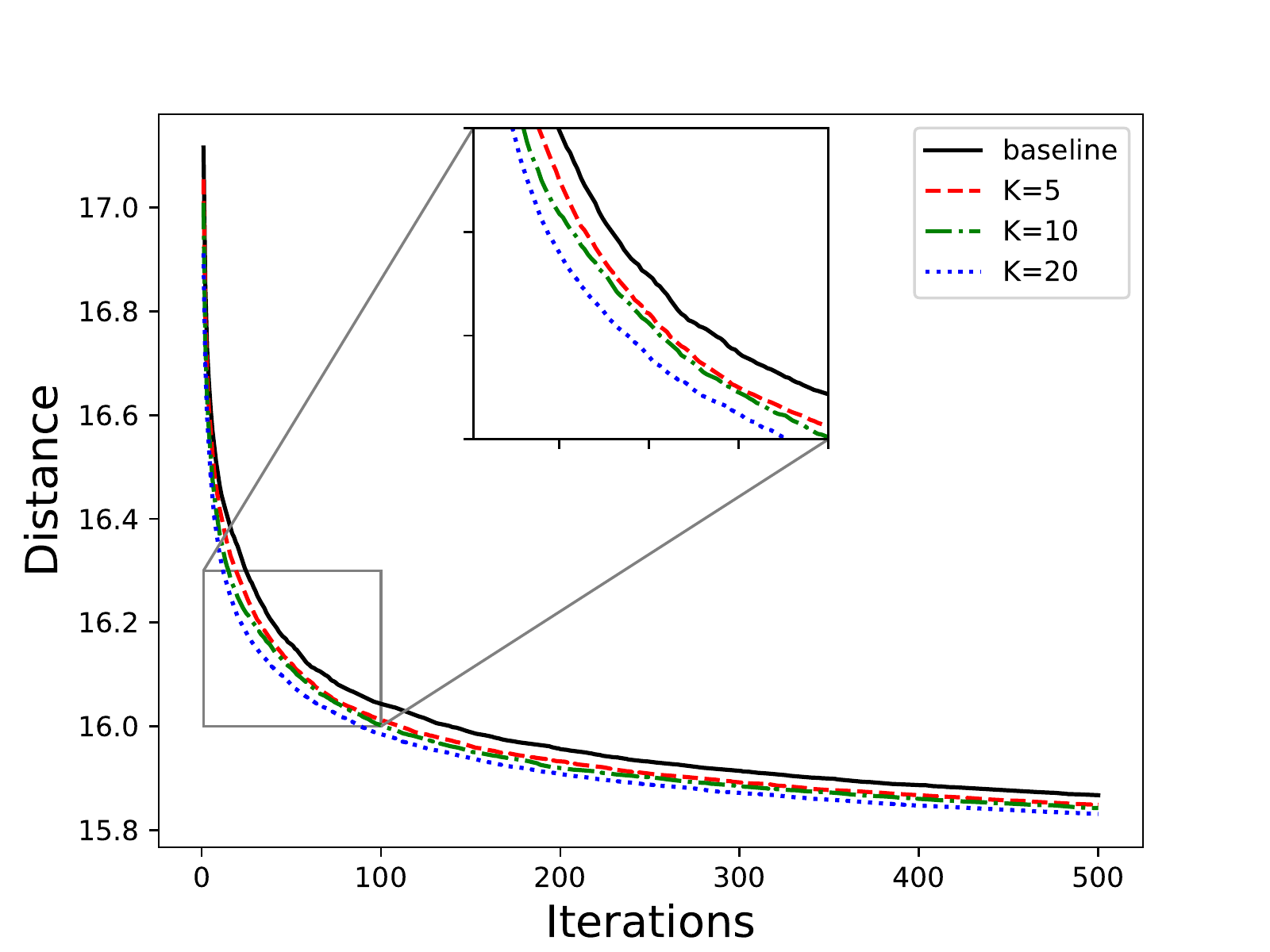} \\
(b) Average best distance after each iteration
\end{tabular}
\caption{CVRP-100: L2R chooses the best offspring.}
\label{FigureN100P20}
\end{figure}

Figure \ref{FigureP20}(a) shows that L2R is able to pick a more promising individual for the BOA to work on. The average distance achieved after each iteration is generally smaller for large $K$ values. However, as shown in Figure \ref{FigureP20}(b), it turns out that for CVRP-20 L2R has no impact on the average best distance over the past iterations, probably because a population of 20 suffices for these relatively small test cases. For CVRP-50 and CVRP-100, Figure \ref{FigureN50P20} and \ref{FigureN100P20} show that L2R improves the average best distance as well.

\begin{figure}[ht]
\centering
\begin{tabular}{c}
\includegraphics[width=0.9\linewidth]{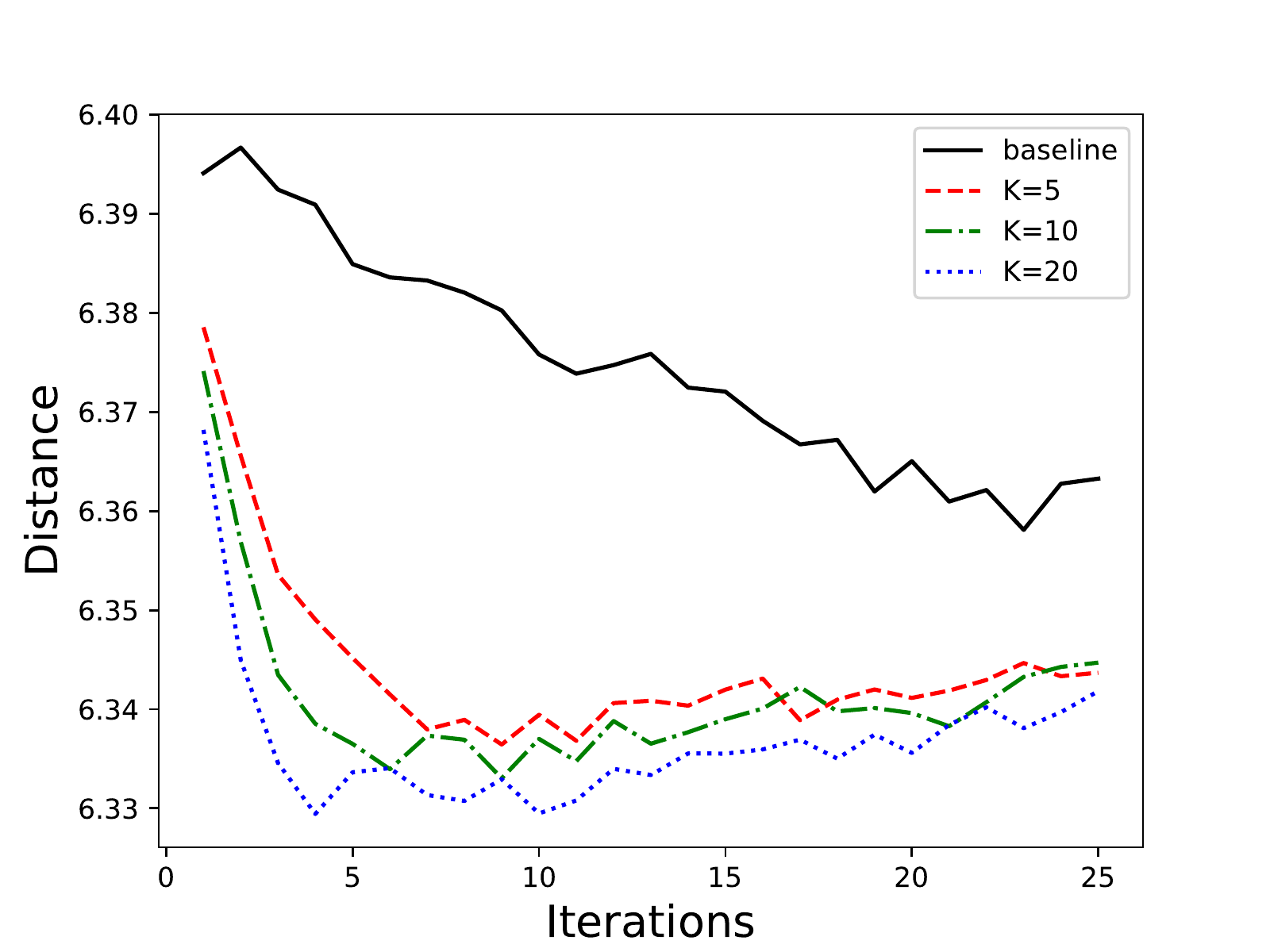}\\
(a) Average distance for each iteration  \\
\includegraphics[width=0.9\linewidth]{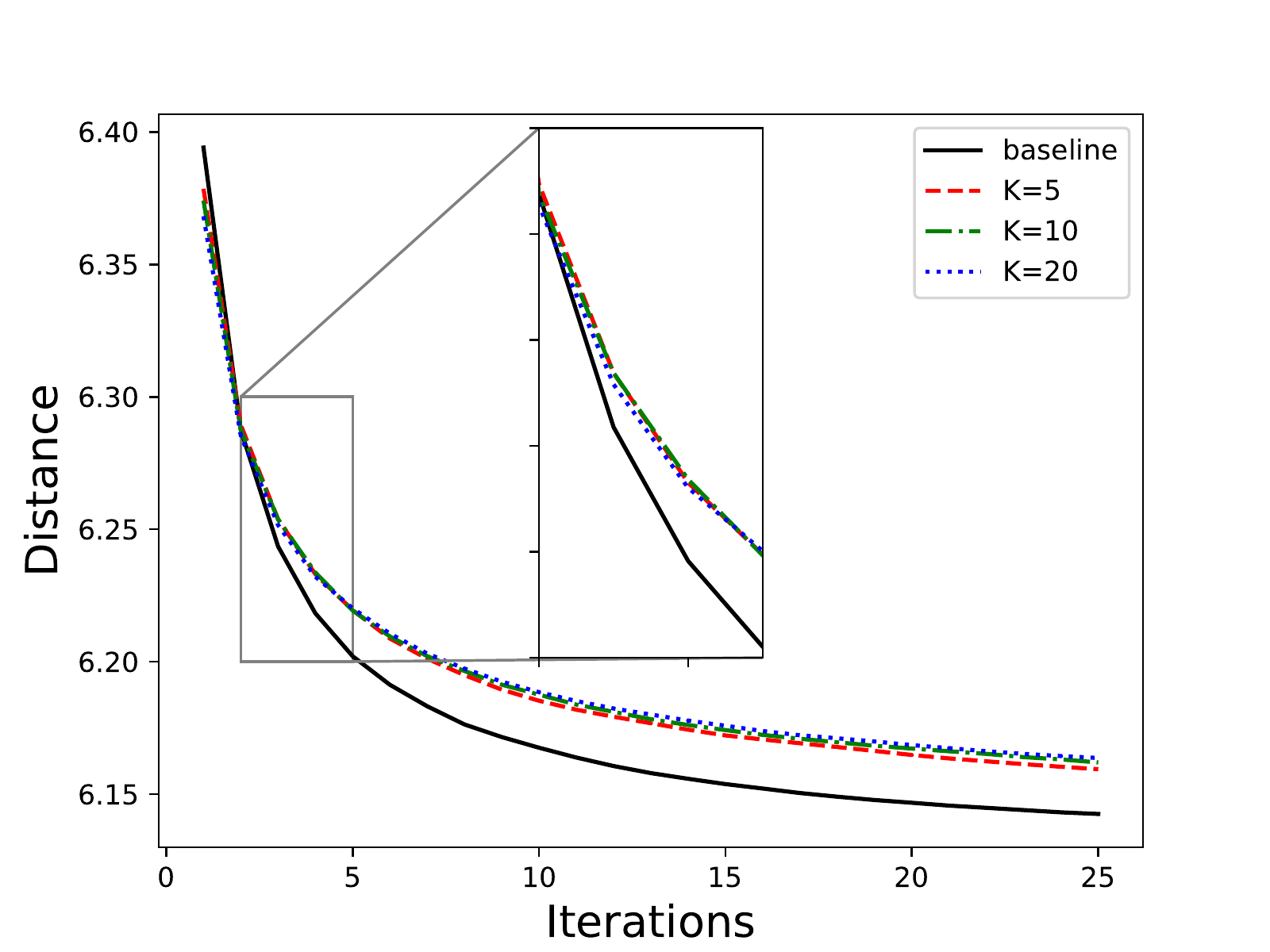} \\
(b) Average best distance after each iteration
\end{tabular}
\caption{CVRP-20: L2R chooses top 20 individuals.}
\label{FigureP20O}
\end{figure}

We also experimented with different ways of using L2R in a population setting. For example, instead of always replacing an individual by its offspring, we generated $K$ offspring, mixed them with the parent generation, and retained top 20 out of these $19 + K$ individuals. It is worthwhile to point out, as shown in Figure \ref{FigureP20O}(b), that the average best distance is worse off as $K$ increases. For a larger $K$, perhaps the population of individuals is more likely to concentrate on offspring of closely related ancestors. Loss of diversity would then lead to the deterioration of the solution quality.

\bibliographystyle{plain}
\bibliography{neurips_2020}